\documentclass[letterpaper, 10 pt, conference]{ieeeconf}  

\IEEEoverridecommandlockouts                              

\usepackage{graphicx}
\usepackage{xcolor}
\usepackage{multirow}
\usepackage{amssymb}
\usepackage{siunitx}
\definecolor{newyellow}{rgb}{1.0, 0.75, 0.0}
\definecolor{newgreen}{rgb}{0.0, 0.5, 0.0}

\usepackage{hyperref}
\usepackage{caption}
\usepackage{bbding}
\usepackage{siunitx}
\usepackage{soul}
\usepackage{makecell}
\usepackage{cancel}
\usepackage{bm}

\newcommand{\strongemph}[1]{\textbf{\textit{#1}}}

\title{RoboHanger: Learning Generalizable Robotic Hanger Insertion\\ for Diverse Garments}

\author{Yuxing Chen$^{1,2}$, Songlin Wei$^{1,2}$, Bowen Xiao$^{1}$, Jiangran Lyu$^{1,2}$, Jiayi Chen$^{1,2}$, Feng Zhu$^{2}$ and He Wang$^{1,2,3\dagger}$
\thanks{$^{1}$CFCS, School of Computer Science, Peking University. $^{2}$Galbot.}%
\thanks{$^{3}$Beijing Academy of Artificial Intelligence.}%
\thanks{$\dagger$Corresponding author: hewang@pku.edu.cn}%
}

\makeatletter
\let\@oldmaketitle\@maketitle
\renewcommand{\@maketitle}{\@oldmaketitle 
  \centering
  \includegraphics[width=0.9\linewidth]{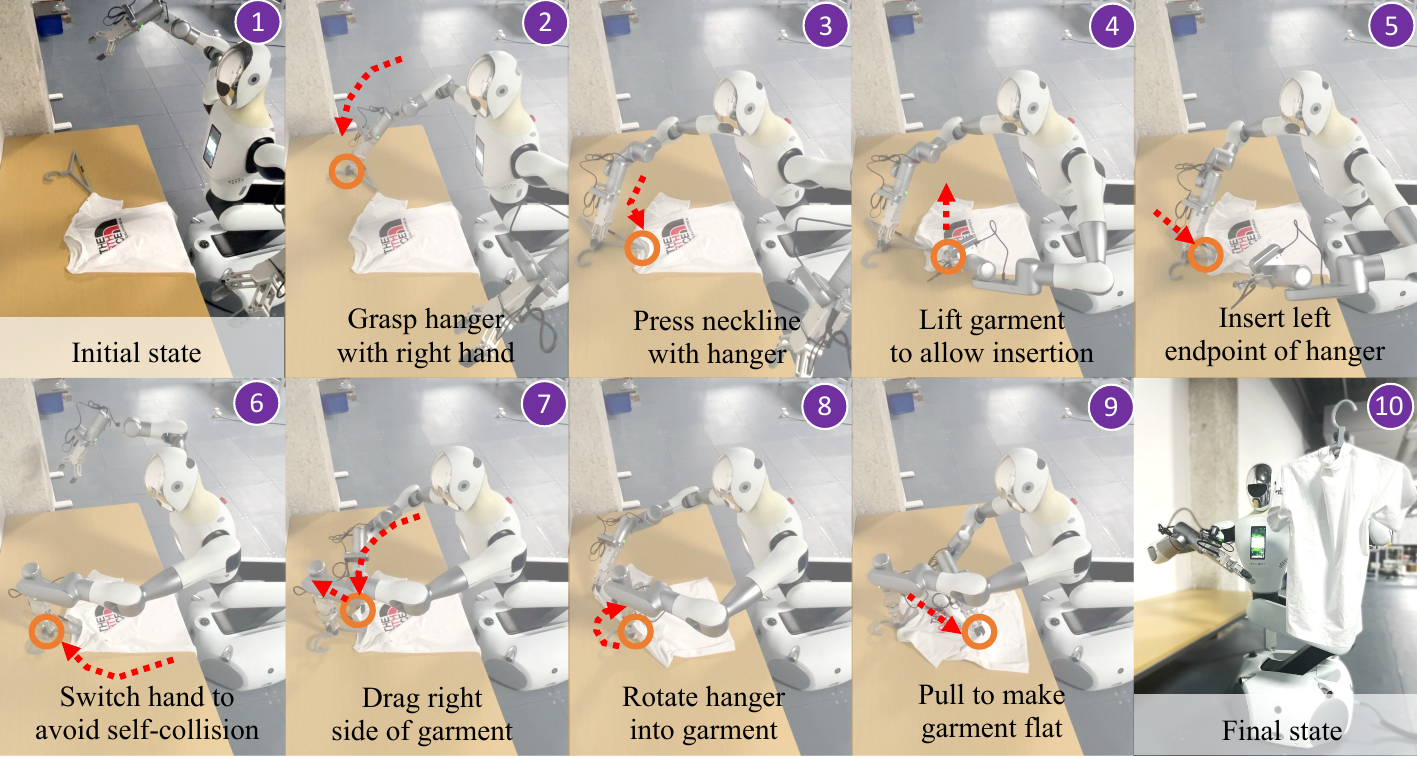}
  \captionof{figure}{\textbf{RoboHanger}: We use a dual-arm robot, where each arm has 7 degrees of freedom (DoF) and is equipped with parallel grippers. A camera is mounted on the robot’s head. Our method is based on visual input, enabling the robot to insert a hanger into the necklines of various garments. 
  }
  \label{fig:teaser}
}
\makeatother

\begin{document}

\maketitle
\addtocounter{figure}{-1}

\thispagestyle{empty}
\pagestyle{empty}


\begin{abstract}

For the task of hanging clothes, learning how to insert a hanger into a garment is a crucial step, but has rarely been explored in robotics. In this work, we address the problem of inserting a hanger into various unseen garments that are initially laid flat on a table. This task is challenging due to its long-horizon nature, the high degrees of freedom of the garments and the lack of data. To simplify the learning process, we first propose breaking the task into several subtasks. Then, we formulate each subtask as a policy learning problem and propose a low-dimensional action parameterization. To overcome the challenge of limited data, we build our own simulator and create 144 synthetic clothing assets to effectively collect high-quality training data. Our approach uses single-view depth images and object masks as input, which mitigates the Sim2Real appearance gap and achieves high generalization capabilities for new garments. Extensive experiments in both simulation and reality validate our proposed method. By training on various garments in the simulator, our method achieves a 75\% success rate with 8 different unseen garments in the real world.

\end{abstract}

\section{Introduction}\label{sec:intro}

As a common household task, using a hanger to hang clothes is a challenging deformable manipulation problem for robots. It involves complex contact dynamics between the hanger and the clothes and requires precise coordination between two robotic arms. The entire task consists of flattening the clothes, inserting the hanger, and hanging the hanger on a rack. Researchers have explored methods to flatten clothes from arbitrary initial configurations~\cite{avigal2022speedfolding, canberk2023cloth}, as well as how to hang objects on racks or hooks~\cite{you2021omnihang, matas2018sim, twardon2015interaction}. However, we have identified that one crucial step—inserting the hanger into the garment—has rarely been studied.

In recent years, learning-based approaches for deformable object manipulation tasks have made significant progress. Zhao et al.~\cite{zhao2024alohaunleashedsimplerecipe} employ end-to-end imitation learning and collect data through teleoperation in the real world to accomplish this task. However, this approach requires substantial human labor for data collection during the training process. 

In this work, we investigate how to use a dual-arm robot to insert a hanger into various unseen garments, as illustrated in Figure \ref{fig:teaser}, employing a Sim2Real approach and without any human labeling effort. To focus on the hanger insertion process, the task begins with the target garment laid flat on the table, a scenario that has been addressed in several previous works~\cite{avigal2022speedfolding, canberk2023cloth}. This task faces three significant challenges: (1) task complexity: the action space is large and it is difficult to define a good reward function, which makes it hard to find a good policy through sampling by directly applying reinforcement learning algorithms; (2) data scarcity: existing simulators lack the necessary physical accuracy and diversity in terms of suitable garments and hangers; and (3) generalizability: garments vary widely in color, shape, and size, making generalization across different garments a non-trivial problem. 

To address the three aforementioned challenges, we propose a method that decomposes the task into two subtasks: inserting the left end of the hanger and inserting the right end of the hanger. Each subtask is accomplished using an action primitive~\cite{bahety2023bag, lyuscissorbot, xue2023unifoldingsampleefficientscalablegeneralizable}, which is a simple trajectory parameterized by two 2D keypoints. These four 2D keypoints are learned through four UNet \cite{ronneberger2015u} models. This low-dimensional parameterization facilitates the learning process and ensures robustness during policy execution in both simulation and reality.

To train the models, we build a simulator and generate over 1,000 hangers and 100 garments for data collection. Our simulation environment is built upon Taichi~\cite{hu2019taichi}, which supports large-scale parallel simulations on GPUs. We generate training data through trial and error within the simulator, minimizing the need for human labeling effort. 

We choose depth and the object's mask as input instead of raw RGB data to mitigate generalization issues. Depth images provide the necessary geometric information for the policies and are naturally unaffected by the varying colors and patterns of different garments. 

Extensive experiments in simulation and the real world validate our proposed method. Trained on 120 different garments in simulation, our method can achieve a final success rate of 75\% on 8 different unseen garments in reality without any fine-tuning with real data. 

\section{Related Work}\label{sec:related}

\subsection{Learning-Based Cloth Manipulation}

Learning-based robotic cloth manipulation policies have recently gained wide interest. Many works collect data from human demonstrations~\cite{zhao2024alohaunleashedsimplerecipe,xue2023unifoldingsampleefficientscalablegeneralizable,peng2024tiebotlearningknottie,avigal2022speedfolding}. The advantage of these methods is the absence of a domain gap, but they require a large amount of manual effort and often suffer from limited generalization. Sim-to-real transfer is a safe and efficient approach~\cite{canberk2023cloth,blanco2024benchmark,zheng2024diffcp,COLTRARO2025105993}. However, previous tasks typically focus only on manipulating the garment itself, without requiring the manipulation of objects other than the garment, such as folding~\cite{longhini2024adafold,mo2023foldsformer,lips2024learningkeypoints} and unfolding~\cite{lin2022learningvisibleconnectivitydynamics,canberk2023cloth,Wu_2024_CVPR}. These tasks require a lower diversity of actions, making it easier for the robot to learn.

In contrast, inserting a hanger into clothes is a long-horizon task that requires the manipulation of both hanger and garment: first, the hanger must be grasped, then both endpoints of the hanger must be inserted into the clothing, and finally, the hanger must be lifted. To efficiently generate a large amount of high-quality training data, we design action primitives to reduce the action space. We build our own cloth simulator to automatically generate high-quality training data and labels through trial and error in simulation.  

\subsection{Robotic Hanging}

Robotic hanging is a task of significant practical importance, yet it has rarely been studied. Previous works have mainly focused on hanging objects on racks or hooks~\cite{twardon2015interaction,kuo2024skthang,Chen_2023}, without using hangers for clothing. In these tasks, typically, only contact between stationary rigid bodies and clothing is involved. However, in our task, the robot needs to simultaneously manipulate both the hanger and the clothing, increasing the complexity of the task. 

Researchers~\cite{Koishihara2017hang} have also tried using a rule-based method to manipulate the hanger, but the method they use lacks sufficient generalizability. Additionally, the manipulation policy requires the hanger and clothing to be pre-attached to the robotic arm's end effector. In our task setup, both the garment and the hanger are initially placed on the table, and the grasping pose is output by our policy. Our manipulation policy for robotic hanger insertion demonstrates strong generalizability across different types of garments.  



\begin{figure}[h]
    \centering
    \includegraphics[width=0.7\linewidth]{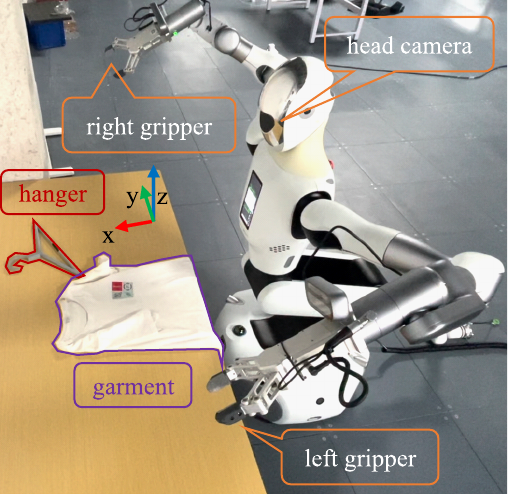}
    \caption{\textbf{Robot system in the real world.} The dual-arm robot is equipped with parallel grippers on both arms and uses a head-mounted camera for observation. The robot's waist can be controlled to lean forward. 
    }
    \label{fig:real_robot}
    \vspace{-4mm}
\end{figure}

\section{Method}

\begin{figure*}[t]
    \centering
    \vspace{3mm} 
    \includegraphics[width=2.0\columnwidth]{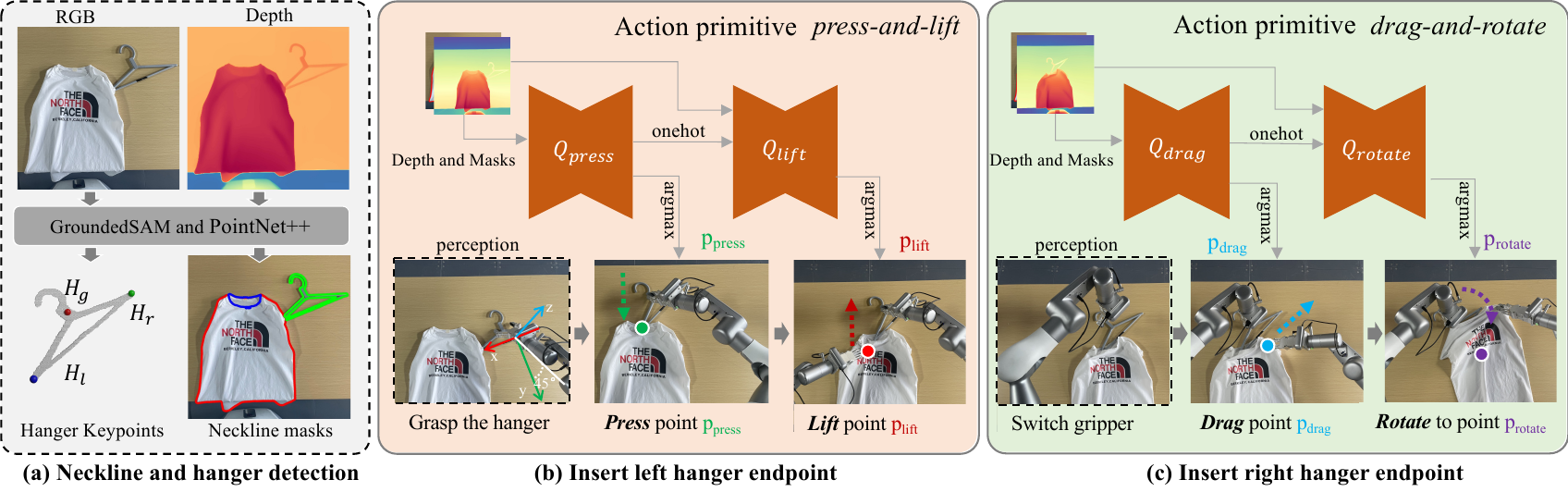}
    \caption{
    \textbf{System Overview.} (a) Before each action primitive, our system takes RGB-D observations as input and segments the hanger, the garment, and its neckline. We pre-detect three keypoints of the hanger at the beginning of the policy. (b) The action primitive \textit{press-and-lift} inserts the \textit{left} endpoint of the hanger into the target garment. (c) The action primitive \textit{drag-and-rotate} inserts the \textit{right} endpoint of the hanger into the target garment. In (b) and (c), four U-Nets ($Q_{press}, Q_{lift}, Q_{drag}, Q_{rotate}$) all take the depth map and masks as input and output 2D value maps indicating the subtask success rate of applying actions at each pixel. We apply \textit{argmax} to obtain a final single point. 
    }
    \label{fig:overview}
    \vspace{-4mm}
\end{figure*}

We focus on the task of inserting a hanger into a garment. As illustrated in Figure \ref{fig:teaser}, the initial state consists of a hanger and a garment placed separately on the table in front of the robot. The goal is to control a dual-arm robot to insert the hanger into the garment through the neckline. The final phase involves the robot lifting the hanger, with the garment successfully hanging. To focus specifically on the task of inserting the hanger, we use the method of previous work~\cite{canberk2023cloth} to initialize the garment. After inserting the hanger into the clothes, we also omit the step of hanging the hanger on a rack. The robot is manually positioned in front of the table, and its base remains stationary during execution. Each of the 7-DoF arms of the robot is equipped with parallel grippers, and a depth camera is mounted on its head for perception, as shown in Figure \ref{fig:real_robot}. 



We will illustrate the overall pipeline in Section \ref{ssec:pipeline} and describe each module, including hanger detection in Section \ref{ssec:perception}, hanger insertion policy in Section \ref{ssec:hanger-insertion}, success criteria in Section \ref{ssec:success_criteria}, action primitive networks in Section \ref{ssec:action_primitives}, the simulation and data collection process in Section \ref{ssec:sim_and_data}, and the whole-body motion controller in Section \ref{ssec:whole-body}. 

\subsection{System Overview}
\label{ssec:pipeline}


Our method takes RGB-D images as input, as illustrated in Figure \ref{fig:overview}. We first detect three keypoints of the hanger: the grasping point $H_g$, the left endpoint $H_l$, and the right endpoint $H_r$. The grasping point indicates where the gripper should grasp the hanger for subsequent motions, while the left and right endpoints mark the locations where the hanger should be inserted into the target garment. Next, we use GroundedSAM~\cite{ren2024grounded} to segment the hanger, garment, and its neckline. The refined depth map, along with the masks, is then concatenated and fed into action primitive networks to predict two critical action primitives: \textit{press-and-lift} and \textit{drag-and-rotate}, which are used to insert the left and right keypoints of the hanger into the garment, respectively. 

\subsection{Garment Segmentation and Hanger Detection}
\label{ssec:perception}
Given a frame of aligned color and depth images, we first use GroundedSAM~\cite{ren2024grounded} to segment and obtain the hanger mask. Using the intrinsic and extrinsic parameters of the camera, we can also obtain the point cloud of the hanger as shown in Figure \ref{fig:overview}(a). We design a simple network consisting of a PointNet++ backbone and a 3-layer MLP detection head that outputs three keypoints $\{H_{g}, H_{l}, H_{r}\}$. To train this network, we synthesize 1,000 meshes of hangers with annotated keypoints. Generating the data and training takes approximately 1 day to run on a single NVIDIA RTX 3090. After predicting these three keypoints, we grasp the hanger using a calculated grasping pose, with the position centered at the point $H_g$ and a rotation of $+45^\circ$ around the vector $(H_l-H_r)$, as shown in the bottom left of Figure \ref{fig:overview}(b). 


\subsection{Hanger Insertion Policy}
\label{ssec:hanger-insertion}

Action primitives~\cite{bahety2023bag,lyuscissorbot,xue2023unifoldingsampleefficientscalablegeneralizable,chen2023autobaglearningopenplastic} are parameterized end-effector motions specifically designed to accomplish complex robotic tasks. They are effective for learning and robust in execution. Our hanger insertion policy consists of two action primitives: \textit{press-and-lift} and \textit{drag-and-rotate}, each responsible for inserting the left and right ends of the hanger into the garment. 

\begin{figure}[h]
    \centering
    \includegraphics[width=\columnwidth]{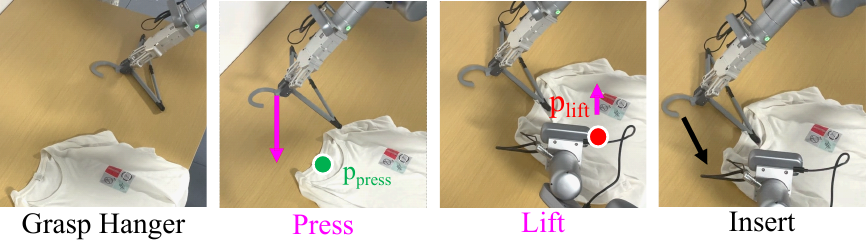}
    \caption{\textbf{Action primitive:} \textit{press-and-lift}. 
    }
    \label{fig:ap1}
\end{figure}

\paragraph{press-and-lift} This action primitive is parametrized by $(p_{press}, p_{lift})=(x_{press}, y_{press}, x_{lift}, y_{lift})\in\mathbb{R}^4$. As illustrated in Figure \ref{fig:ap1}, after successfully grasping the hanger, we first use the left endpoint $H_l$ of the hanger to press the neckline at point $p_{press}$. With the garment pressed by the hanger, we then lift the garment at point $p_{lift}$ with the other hand without moving the garment. Finally, we insert the hanger into the garment at a fixed distance. By combining the \textit{press} and \textit{lift} actions, we can overcome the challenge of the gripper grasping both layers of the garment, which may otherwise hinder the insertion of the hanger. 



\begin{figure}[h]
    \centering
    \vspace{3mm} 
    \includegraphics[width=\columnwidth]{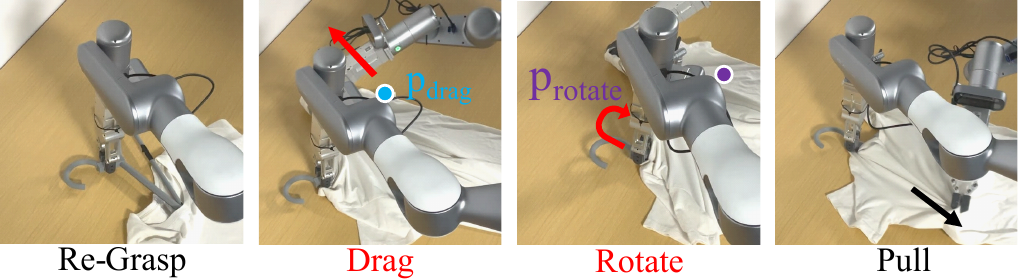}
    \caption{\textbf{Action primitive:} \textit{drag-and-rotate}. 
    }
    \label{fig:ap2}
\end{figure}

\paragraph{drag-and-rotate} This action primitive is parametrized by $(p_{drag}, p_{rotate})=(x_{drag}, y_{drag}, x_{rotate}, y_{rotate})\in\mathbb{R}^4$. As illustrated in Figure \ref{fig:ap2}, we first move the left gripper to the position where the right gripper previously released the hanger and re-grasp the hanger to avoid self-collision between the two arms. Next, we use the right gripper to grasp the garment at point $p_{drag}$ and drag it forward to the right. Then, we use the left gripper to rotate the hanger while moving the right end to $p_{rotate}$ and aligning it to the correct angle. Finally, we slightly pull the garment back to arrange it properly. 

\subsection{Success Criteria} 
\label{ssec:success_criteria}

For each action primitive, the success criterion is whether the hanger's endpoint ($H_l$ for the first action primitive and $H_r$ for the second) is inserted into the garment. Similarly, if both endpoints are inserted, the task is considered completed. We use \( S_1 \) and \( S_2 \) to represent the success status of each action primitive. \(S_1=1\) indicates the successful insertion of the left endpoint, and \(S_2=1\) indicates the successful insertion of the right endpoint. \(S_1=0\) or \(S_2=0\) indicates failure. 

To detect successful insertion during simulation, we cast multiple rays randomly from the endpoint in different directions. If the proportion of rays intersecting with the garment's mesh exceeds a threshold (0.95 for the left endpoint and 0.9 for the right in our experiments), we consider the endpoint to be covered by the garment. This proportion is similar to the generalized winding number~\cite{generalized_winding_numbers} in $\mathbb{R}^{3}$ space. The thresholds are set manually by inspecting many successful and unsuccessful trajectories and meshes in simulation. 


\subsection{Action Primitive Networks}
\label{ssec:action_primitives}

For each action primitive, we utilize two networks to predict the actions of the left and right hands, respectively. In the first action primitive, we use the network $Q_{press}$ to predict the press position $p_{press}$, and the network $Q_{lift}$ to predict the lift position $p_{lift}$. Likewise, for the second action primitive, we use network $Q_{drag}$ to predict the drag position $p_{drag}$, and network $Q_{rotate}$ to predict the rotate position $p_{rotate}$. These networks are similar to action-value functions in reinforcement learning, but we only consider the success rate within a single subtask.

The network architecture is based on UNet~\cite{ronneberger2015u}. The network input, denoted as $O$, consists of a depth image and object masks. Specifically, we use the garment mask and neckline mask when predicting $p_{press}$ and $p_{lift}$ for the \textit{press-and-lift} action, and the garment mask and hanger mask when predicting $Q_{drag}$ and $Q_{rotate}$ for the \textit{drag-and-rotate} action. The network output is an image of the same size as the input, with each pixel representing the success rate of inserting one endpoint of the hanger, as mentioned in \ref{ssec:success_criteria}. 

We employ the following dataflow to collaboratively learn two networks in an action primitive. During inference, $Q_{press}$ predicts $p_{press}$, and then $Q_{lift}$ predicts $p_{lift}$ conditioning on $p_{press}$. During training, we use trained $Q_{lift}$ to provide supervision for $Q_{press}$. Specifically, given data of the form \( (O_1, p_{press}, p_{lift}, S_1) \), where \( O\in\mathbb{R}^{C\times H\times W} \) is the observation, \( p_{press},p_{lift}\in\mathbb{R}^2 \), \( S_1\in\{0,1\} \) denotes success or failure. We use \( Q[y;x] \in \mathbb{R} \) to represent the input image \( x \) to the UNet, and select the \( y \)-th pixel in the output image. The loss function of $Q_{lift}$ is: 
\begin{equation}
\mathcal{L}_{lift}=BCE(Q_{lift}[p_{lift};O_1,p_{press}],S_1)
\end{equation}

where $BCE$ denotes the binary cross-entropy loss. The input \( O_1'=(O_1,p_{press})\in\mathbb{R}^{(C+1)\times H\times W}\) to the network \( Q_{lift} \) is modified to be \( O_1 \) concatenated with the one-hot map of \( p_{press} \), which is used to represent the conditioning on \( p_{press} \). The maximum value of the $Q_{lift}$'s output can serve as supervision for $Q_{press}$: 
\begin{equation}
\hat{S_1}=\max\{Q_{lift}[\cdot;O_1,p_{press}]\}>0.5
\end{equation}
\begin{equation}
\mathcal{L}_{press}=BCE(Q_{press}[p_{press};O_1],\hat{S_1})
\end{equation}

Similarly, given data of the form \( (O_2, p_{drag}, p_{rotate}, S_2) \), the loss functions of $Q_{drag}$ and $Q_{rotate}$ are: 

\begin{equation}
\mathcal{L}_{rotate}=BCE(Q_{rotate}[p_{rotate};O_2,p_{drag}],S_2)
\end{equation}
\begin{equation}
\hat{S_2}=\max\{Q_{rotate}[\cdot;O_2,p_{drag}]\}>0.5
\end{equation}
\begin{equation}
\mathcal{L}_{drag}=BCE(Q_{drag}[p_{drag};O_2],\hat{S_2})
\end{equation}

We apply \textit{argmax} to the resulting 2D value map to obtain a single final pixel. When executing these action primitives, we back-project pixels into the world frame. The parameters of the action primitives only use the x and y coordinates, while the z coordinate is fixed. This is to avoid failures caused by depth camera inaccuracies.

\subsection{Data Collection}
\label{ssec:sim_and_data}

\paragraph{Simulation environment and assets} We observe that existing simulators, such as PyFlex~\cite{li2018learning} and MuJoCo~\cite{mujoco2012}, either lack the accuracy or efficiency needed for our task. The main physics accuracy issue is that they do not handle collisions and dynamics of clothing very well. Therefore, we build our simulation environment based on Taichi~\cite{hu2019taichi}. 

We use the finite element method (FEM) and implicit time integration with a single-step Newton iteration~\cite{baraff2023large} for the physical simulation. Cloth self-collision and cloth-to-rigid-body collision are handled with a cubic energy function. Cloth self-intersection is prevented using Continuous Collision Detection (CCD). The collision force between cloth and rigid bodies is computed using the Signed Distance Function (SDF) of the latter. We do not perform simulation of dynamics or collision handling for rigid and articulated bodies; instead, we compute the position and velocity of objects using kinematic relationships. Depth images are rendered with SAPIEN~\cite{Xiang_2020_SAPIEN}, and we design the 3D models of clothes and hangers heuristically. 


\paragraph{Data collection strategy} We use an iterative data collection approach. Specifically, we first design a state-based heuristic policy to collect trajectories and use this dataset to initialize our networks. This policy achieves a success rate of approximately 50\%, generating a balanced dataset with both positive and negative samples. The heuristic policy attempts to find actions that satisfy all of our predefined rules through sampling. 


Next, we use the network policy to collect more data in simulation, incorporate this online data into the dataset, and iteratively fine-tune the network. During the initialization phase, we collect 12,000 trajectories. In the subsequent fine-tuning phase, we collect 1,200 trajectories per iteration, with a total of 5 iterations, resulting in 18,000 trajectories in total. Simulation and rendering take approximately 360 GPU hours on an NVIDIA RTX 3090, and training the network requires around 1 hour.



\subsection{Whole-body Motion Controller} 
\label{ssec:whole-body}
Since our hanger insertion task requires a large working space — with some actions necessitating the agent to lean forward without colliding with the table — we implement a custom whole-body controller to ensure the action space does not become a bottleneck. Specifically, after specifying the position of the robotic arm's end effector, we only constrain the grippers' fingertips to remain parallel to the table, leaving the other two rotational degrees of freedom unrestricted. Additionally, for steps that require an even larger workspace, we adjust the waist joint angles to allow the robot to lean forward. These angles are computed using a linear function based on the target positions of the left and right end effectors. 

\subsection{Sim-to-real Transfer} 
\label{ssec:sim2real}

For sim-to-real visual gap, we randomize the camera's intrinsic and extrinsic parameters, depth maps, and masks during network training. This noise enables the network to handle a wider range of input distributions. When deploying in the real world, we use the method from~\cite{wei2024d3romadisparitydiffusionbaseddepth} to obtain a refined depth map. For sim-to-real physics gap, we randomize the Young's modulus and the density of the clothing, as well as the coefficient of friction of the garment in the simulation. 

The head camera is carefully calibrated. Apart from this, we do not perform any additional real fine-tuning and directly transfer the policy learned in simulation to the real world.

\section{Experiment}\label{sec:experiment}

\subsection{Metrics}

We evaluate the performance of the policy based on the success rate \(S\) of hanger insertion. \( S \) can be further divided into: the success rate of inserting the left endpoint, \( S_1 \), and the success rate of inserting the right endpoint after successfully inserting the left endpoint, \( S_2 \). In our policy, the left end is always inserted first, followed by the right end. Therefore, $S=S_1 \times S_2$. The success rate is computed based on 720 trials in simulation and 16 trials in the real world.


\begin{figure*}[t]
    \centering
    \vspace{3mm} 
    \includegraphics[width=2\columnwidth]{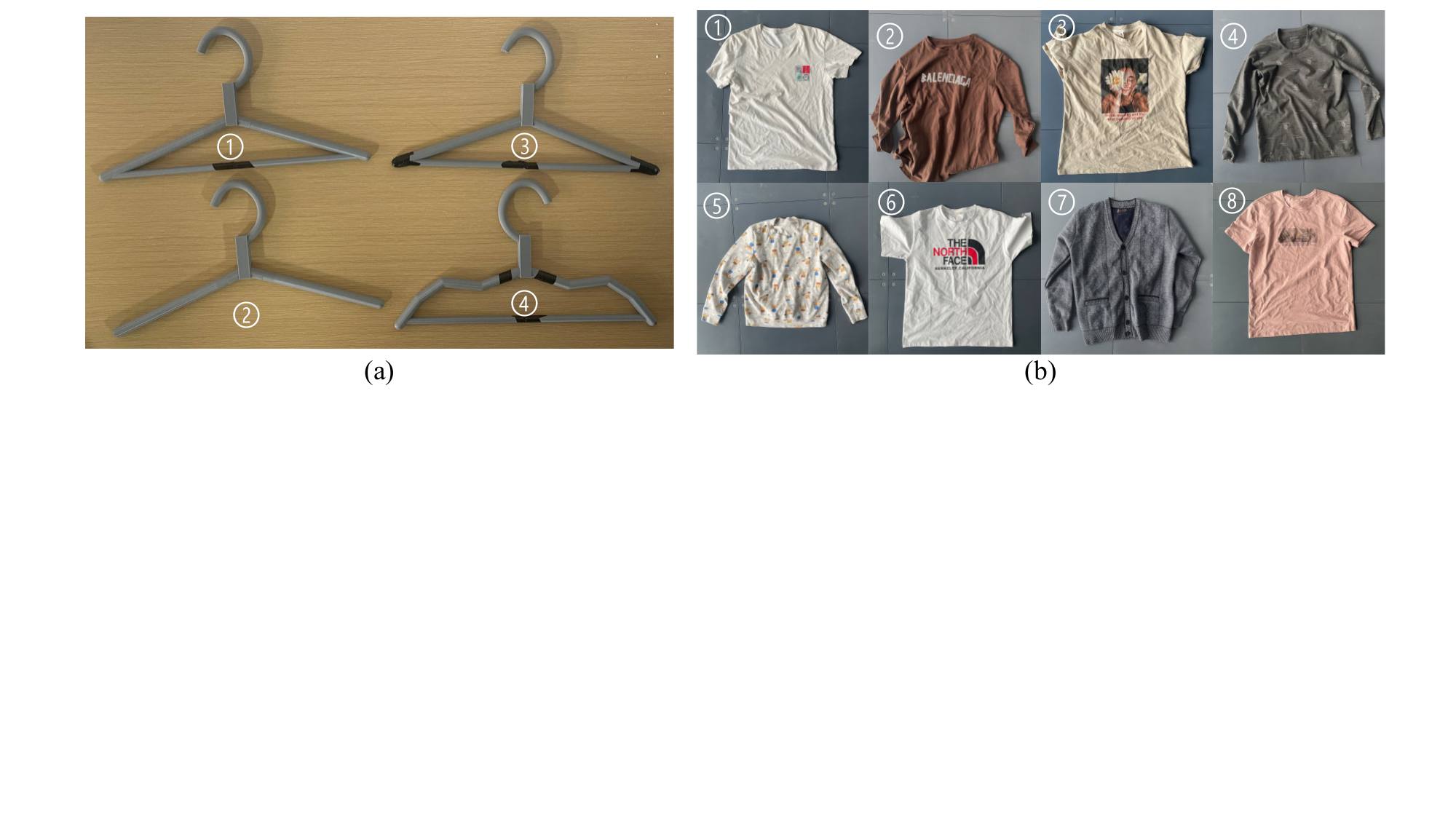}
    \vspace{-50mm}
    \caption{\textbf{Real world assets.} In the real world, the hangers and garments used for evaluation are as follows: Figure (a) shows the hangers we use, each with a width of approximately \SI{40}{\centi\meter}. These hangers vary in shape and coefficient of friction. Figure (b) shows our test garments, which differ in \strongemph{color}, \strongemph{thickness}, \strongemph{size}, \strongemph{neckline shape}, and \strongemph{coefficient of friction}. 
    }
    \label{fig:real asset}
\end{figure*}


\subsection{Comparison with Baselines in Simulation}

We first compare our method with several state-of-the-art reinforcement learning and imitation learning approaches in simulation. Among the 144 synthetic clothing meshes, we randomly divided them into 120 training garments and 24 testing garments. The hangers used for training and testing are the ones shown in Figure \ref{fig:real asset} (a.1) and (a.2). 



\begin{table}[h]
    \centering
    \begin{tabular}{|l||c|c|c|c|c|}
    \hline 
        Methods & Input & Output & S1(\%)~$\uparrow$ & S2(\%)~$\uparrow$ & S(\%)~$\uparrow$ \\
    \hline 
        Ours & V & Prim & $88.7$ & $\mathbf{95.9}$ & $\mathbf{85.1}$ \\
    \hline
        Heuristic & S & Prim & $59.1$ & $75.4$ & $44.5$ \\
    \hline
        IL & V & Prim & $85.8$ & $91.7$ & $78.7$ \\
    \hline
        SAC~\cite{haarnoja2018softactorcriticoffpolicymaximum} & V & Prim & $37.9$ & $56.0$ & $21.2$ \\
    \hline
        ACT~\cite{zhao2023learningfinegrainedbimanualmanipulation}& V & Joint & $\mathbf{94.8}$ & $67.0$ & $63.5$ \\
    \hline
        DP~\cite{chi2024diffusionpolicyvisuomotorpolicy} & V & Joint & $82.8$ & $59.7$ & $49.5$ \\
    \hline
    \end{tabular}
    \caption{Comparison with baseline policies in simulation. 
    }
    \label{tab:sim baseline}
\end{table}

We present the comparisons in Table \ref{tab:sim baseline}. \textit{Input} refers to the type of input used by the policy, which can be either visual input (V) or state input (S). \textit{Output} refers to the type of output generated by the policy, which can either be action primitives (Prim) or joint angles (Joint). Our method performs the best in simulation, achieving an overall success rate of 85.1\%. 

\textit{Heuristic} refers to a heuristic algorithm that we manually designed based on the current mesh of the garment, as described in \ref{ssec:sim_and_data} (c). Since these rules are manually defined, their performance is not as good as methods that learn good actions from large amounts of data based on action labels.

\textit{IL} stands for imitation learning, which is also based on action primitives and UNets. We treat the task as a multi-class classification task (selecting a pixel in the image) using cross-entropy loss and only retain positive samples for training the UNets. Compared with our method, \textit{IL} cannot utilize failure data for training, which results in degraded performance.


\textit{SAC} stands for Soft Actor-Critic algorithm~\cite{haarnoja2018softactorcriticoffpolicymaximum}. In SAC, the input to the value network consists of both the observation and the action. The action is first transformed into an image where each pixel value represents the inverse of the distance to that action. This action image is then concatenated with the observation and fed into a CNN. The action network is also a CNN that directly outputs \((X_i, Y_i)\). Compared to our method, SAC requires learning an additional action network, while our method directly finds the maximum position on the output of the value network. The success rate is lower because the additional action network causes additional training difficulty.

\textit{ACT} and \textit{DP} (diffusion policy) are methods proposed by~\cite{zhao2023learningfinegrainedbimanualmanipulation} and~\cite{chi2024diffusionpolicyvisuomotorpolicy}, respectively. We directly use data generated by heuristic policy to train an end-to-end imitation learning policy. The network's input consists of raw depth images from the head camera, along with the masks of the clothing and the hanger. The output of the network is the angle of each robot joint. We only retain successful trajectories from the data for training, with a total of approximately 10,000 trajectories. All trajectories begin with the hanger already being grasped. Compared to our output of only 4 actions, these end-to-end policies need to output about 200 actions in a complete trajectory. These policies perform excellently when inserting the first endpoint, but due to accumulated errors and lack of data to recover from errors, they struggle with inserting the second endpoint. 

\subsection{Ablation Studies in Simulation}

We provide the ablation results in Table \ref{tab:sim ablation} to validate our design choices. 

\begin{table}[h]
    \centering
    \begin{tabular}{|l||c|c|c|}
    \hline 
        & S1(\%)~$\uparrow$ & S2(\%)~$\uparrow$ & S(\%)~$\uparrow$ \\
    \hline 
        Ours & $\mathbf{88.7}$ & $\mathbf{95.9}$ & $\mathbf{85.1}$ \\
    \hline
        w/o online data & $86.5$ & $94.7$ & $81.8$ \\
    \hline
        w/o neckline mask & $85.6$ & $94.2$ & $80.6$ \\
    \hline
    \end{tabular}
    \caption{Ablation study in simulation.}
    \label{tab:sim ablation}
\end{table}

\textit{w/o online data} refers to training the network using only offline data (generated by the heuristic policy), without incorporating online data (generated by the UNet policy), while keeping the data quantity the same. Fine-tuning with online data results in an improvement in the final success rate. \textit{w/o neckline mask} indicates training the policy without using the neckline mask. It is evident that the neckline mask also contributes to an increase in the final success rate. 

\subsection{Comparison with Baselines in Real World}

\begin{figure*}[t]
    \centering
    \vspace{3mm} 
    \includegraphics[width=1.9\columnwidth]{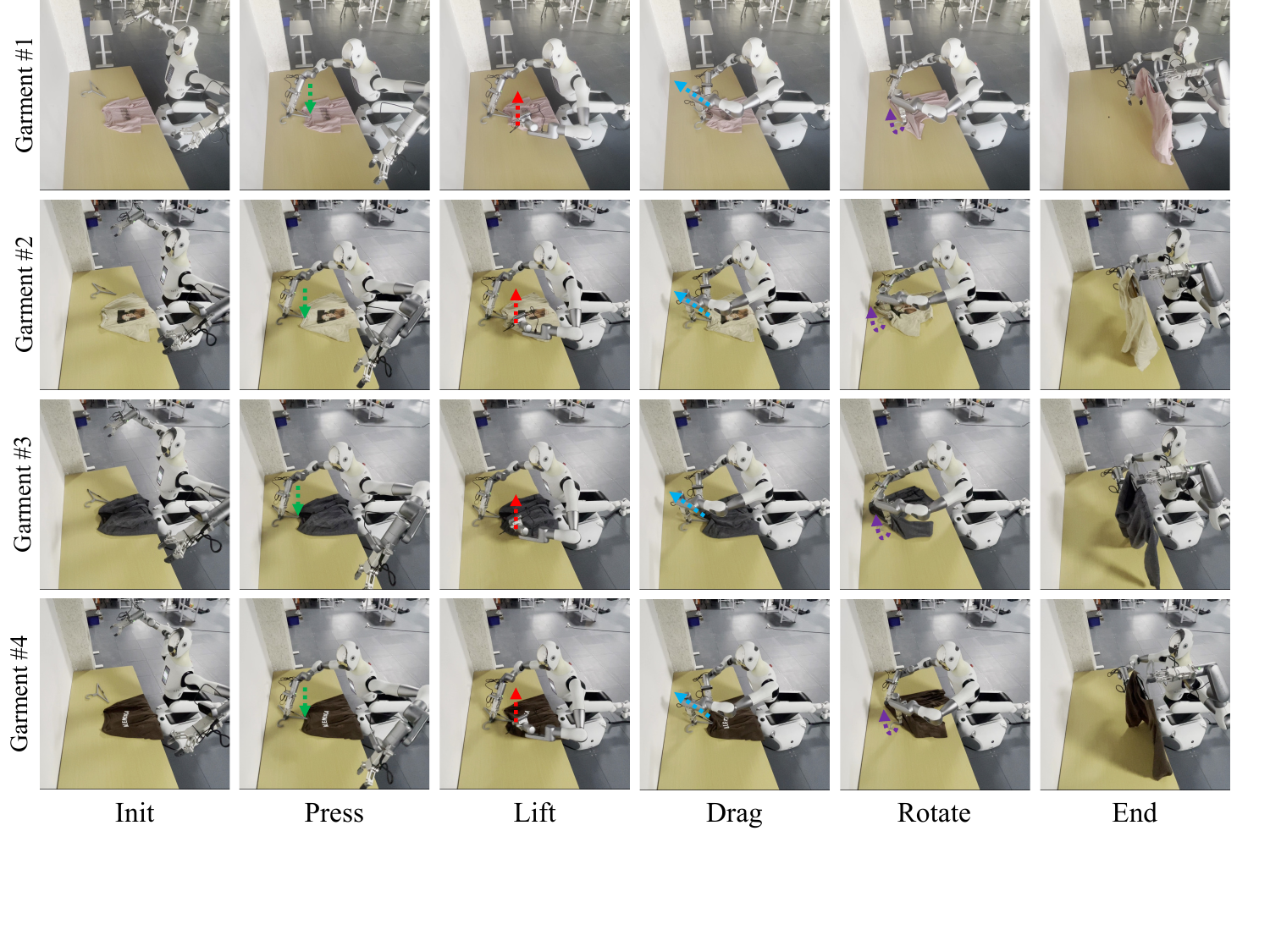}
    \vspace{-15mm}
    \caption{The figure shows the different garment states after applying each action primitive of RoboHanger system in the real world.}
    \label{fig:qualitative}
\end{figure*}

In real world experiments, we test 8 different garments and \textit{Hanger 1}, as shown in Figure \ref{fig:real asset}. For a fair comparison, we use the robot to initialize the garment configuration. Additionally, to evaluate our policy's performance under different initializations, we also report the policy's success rate with human initialization. More qualitative results are shown in Figure \ref{fig:qualitative}. 

\begin{table}[h]
    \centering
    \begin{tabular}{|l||c|c|c|c|}
    \hline 
        & Output & S1(\%)~$\uparrow$ & S2(\%)~$\uparrow$ & S(\%)~$\uparrow$ \\
    \hline 
        Ours & Prim & $\mathbf{93.8}$ & $80.0$ & $\mathbf{75.0}$ \\
    \hline
        IL & Prim & $75.0$ & $\mathbf{83.3}$ & $62.5$ \\
    \hline
        ACT & Joint & $12.5$ & $0.0$ & $0.0$ \\
    \hline
        Fix & Prim & $75.0$ & $41.7$ & $31.2$ \\
    \Xhline{1.2pt}
        Ours(H) & Prim & $93.8$ & $85.3$ & $80.0$ \\
    \hline
    \end{tabular}
    \caption{Experiments in the real world.}
    \label{tab:real baseline}
\end{table}

We present the results of our policy compared to other methods on the real robot in Table \ref{tab:real baseline}. We include \textit{IL} (imitation learning), which performs the best among the baseline policies in simulation; \textit{ACT} (a policy that directly outputs joint angles); and \textit{Fix} (a policy that outputs hardcoded parameters for both action primitives) as baselines. Our method outperforms all baseline methods in terms of the final success rate, which demonstrates the efficacy of our approach. 

Although the ACT policy shows a certain level of success in simulation, particularly with a high success rate when inserting the first endpoint, it suffers from a severe sim-to-real gap and rarely succeeds in the real world. 

The fixed policy is a manually tuned policy designed for the test garments. We find that the fixed policy still achieves high success rates for inserting the left endpoint, as the initial position of the garment remains relatively fixed in our experiments. However, when inserting the right endpoint, the success rate of the fixed policy drops significantly due to unforeseen deformations of clothes during insertion of the left endpoint. 

In the last row of the table (\textit{Ours(H)}), we further report the performance of our policy when the garment is initialized by a human. When the garment is initialized by the robot, the success rate shows only a slight decrease compared to human initialization, demonstrating the robustness of our policy to variations in the initialization process. 

\subsection{Generalization of Hangers in Real World}

To further study the generalizability of our method across different hangers, we report the performance of our policy using various hangers in Table \ref{tab:real different hanger}. Here, Hanger 2 is a hanger seen during training but without a crossbar. Hanger 3 and Hanger 1 have the same geometry, but Hanger 3 has a much rougher surface. Hanger 4 is a hanger that is not seen during training. 

\begin{table}[h]
    \centering
    \begin{tabular}{|l||c|c|c|}
    \hline 
        & S1(\%)~$\uparrow$ & S2(\%)~$\uparrow$ & S(\%)~$\uparrow$ \\
    \hline 
        Hanger 1 & $\mathbf{93.8}$ & $\mathbf{80.0}$ & $\mathbf{75.0}$ \\
    \hline
        Hanger 2 & $81.2$ & $23.1$ & $18.8$ \\
    \hline
        Hanger 3 & $75.0$ & $75.0$ & $56.2$ \\
    \hline
        Hanger 4 & $\mathbf{93.8}$ & $66.7$ & $62.5$ \\
    \hline
    \end{tabular}
    \caption{Comparison of performance with different hangers in the real world.}
    \label{tab:real different hanger}
\end{table}

\begin{figure}[h]
    \centering
    \includegraphics[width=1.0\columnwidth]{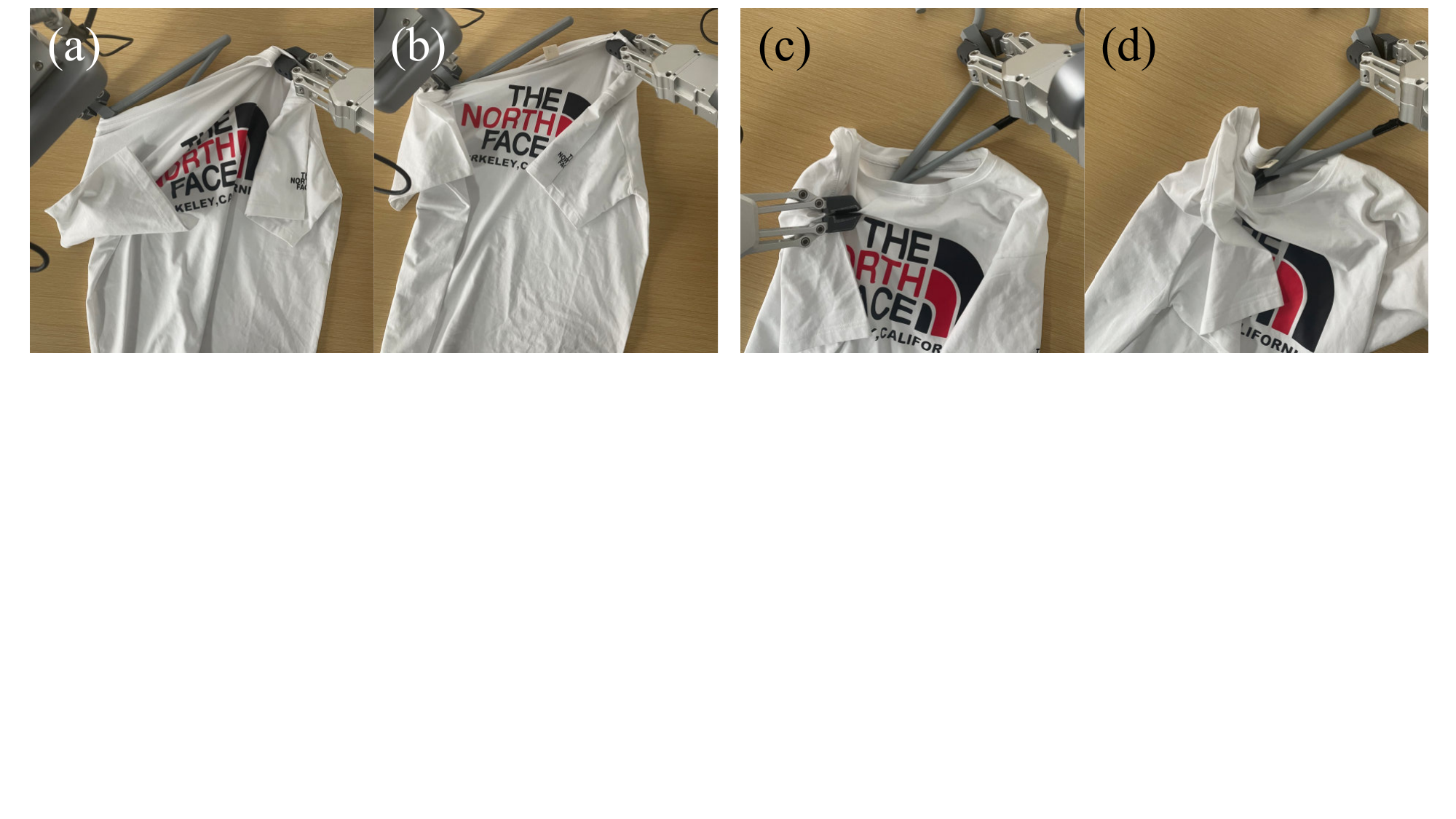}
    \vspace{-30mm}
    \caption{\textbf{Results with different hangers in the real world.} Issues may arise when changing hangers in reality. Figure (a) and Figure (c) show results with Hanger 1. As shown in Figure (b), when switching to Hanger 2, it becomes difficult to separate the upper and lower layers of the clothing while inserting the right end. Figure (d) illustrates that with Hanger 3, which has higher friction, the clothing tends to get wrinkled during the insertion of the left end, causing the garment to become stuck in a worse state. 
    }
    \label{fig:real defferent hangers}
\end{figure}

Although Hanger 2 was seen during training, there is a significant performance drop when inserting the right endpoint. As shown in Figures \ref{fig:real defferent hangers}(a) and \ref{fig:real defferent hangers}(b), these two images compare the state when inserting the right endpoint using different hangers. The presence of a crossbar helps the hanger separate the upper and lower layers of the clothing, making insertion easier. When using Hanger 3, as shown in Figures \ref{fig:real defferent hangers}(c) and \ref{fig:real defferent hangers}(d), if the hanger has too much friction, it may drag the clothes while moving, causing the clothes to become very wrinkled, which in turn reduces the success rate. When using Hanger 4, the complexity of the hanger's shape also affects the final success rate. 

\section{Failure Cases and Limitations}\label{sec:limitation}



\textbf{Failure Cases.} For the insertion of the first endpoint, the main failure mode is the inability to separate the front and back layers of the garment. The primary reason is the complex contact forces between the garment and the hanger. For inserting the second endpoint, the main failure mode is the inability to rotate the hanger into the garment neckline. The primary reason is the large deformation of the garment at this stage, which increases the sim-to-real gap.

\textbf{Limitations.} Currently, garment grasping in the simulation is simplified as the garment being attached to the gripper. This physical modeling limits the exploration of potentially better strategies. A more fine-grained simulation of garment-gripper contact could further improve the overall success rate. For garments with less elasticity, such as shirts, more precise force feedback control may be required to avoid damaging both the hanger and the garment. Additionally, closed-loop control could also be introduced to help recover from failure cases. 

\section{Conclusion}\label{sec:conclusion}

In this work, we explore the task of inserting a hanger into the neckline of an unseen garment. For this challenging task, we propose a simulation environment and a data collection pipeline. We demonstrate the efficiency of training the algorithm in a simulation environment, as well as its effectiveness and generalizability in the real world. Our pipeline enables the efficient collection of demonstrations for future robotic foundation models that require large amounts of diverse data. Future work may involve relaxing the assumptions made for this task and extending the method to handle a broader variety of garments. In addition, exploring the input of RGB images could be an interesting research direction. 

\bibliographystyle{IEEEtran}
\bibliography{IEEEabrv,ref}

\end{document}